\documentclass[wcp]{jmlr}

\usepackage{amsmath,amssymb,graphicx,url}
\usepackage{longtable}

\usepackage{booktabs}
\usepackage[load-configurations=version-1]{siunitx} 
\usepackage{mathrsfs}
\usepackage{adjustbox}
\usepackage{multirow}
\usepackage{xcolor,colortbl}
\usepackage[flushleft]{threeparttable}

\theorembodyfont{\upshape}
\theoremheaderfont{\scshape}
\theorempostheader{:}
\theoremsep{\newline}

\jmlrvolume{204}
\jmlryear{2023}
\jmlrworkshop{Conformal and Probabilistic Prediction with Applications}
\jmlrproceedings{PMLR}{Proceedings of Machine Learning Research}

\title[Approximating Score-based Explanation Techniques Using Conformal Regression]{Approximating Score-based Explanation Techniques \\Using Conformal Regression}

    \author{\Name{Amr Alkhatib} \Email{alkhat@kth.se}\\
   \Name{Henrik Boström} \Email{bostromh@kth.se}\\
   \Name{Sofiane Ennadir} \Email{ennadir@kth.se}\\
   \addr School of Electrical Engineering and Computer Science, KTH Royal Institute of Technology, Sweden\\
   \Name{Ulf Johansson} \Email{ulf.johansson@ju.se}\\
   \addr Dept. of Computing, Jönköping University, Sweden}

\editor{Harris Papadopoulos, Khuong An Nguyen, Henrik Boström and Lars Carlsson}

\begin{document}

\maketitle

\begin{abstract}
Score-based explainable machine-learning techniques are often used to understand the logic behind black-box models. However, such explanation techniques are often computationally expensive, which limits their application in time-critical contexts. Therefore, we propose and investigate the use of computationally less costly regression models for approximating the output of score-based explanation techniques, such as SHAP. Moreover, validity guarantees for the approximated values are provided by the employed inductive conformal prediction framework. We propose several non-conformity measures designed to take the difficulty of approximating the explanations into account while keeping the computational cost low. We present results from a large-scale empirical investigation, in which the approximate explanations generated by our proposed models are evaluated with respect to efficiency (interval size). The results indicate that the proposed method can significantly improve execution time compared to the fast version of SHAP, TreeSHAP. The results also suggest that the proposed method can produce tight intervals, while providing validity guarantees. Moreover, the proposed approach allows for comparing explanations of different approximation methods and selecting a method based on how informative (tight) are the predicted intervals.
\end{abstract}
\begin{keywords}
Inductive conformal prediction $\cdot$ Explainable machine learning $\cdot$ Multi-target regression 
\end{keywords}

\section{Introduction}
\label{sec:intro}

Explainable machine learning has become increasingly important as machine learning algorithms are applied to real-world problems within various domains, including healthcare, finance, and criminal justice, among others \citep{LakkarajuKCL17}. In many cases, it is essential that the output of such machine learning algorithms are transparent and can be understood by the users. 

Explanation methods are classified based on the scope of the explanations they produce, which can be either local or global \citep{molnar2022}. Local explanation methods provide instance-based explanations for a single prediction of a black-box model, while global explanation methods provide an overview of how a model behaves in general \citep{molnar2022}.
Explanation methods can also be classified into two main categories: model-agnostic and model-specific. Model-agnostic techniques are designed to explain any black-box model and have been described in previous research, e.g., \cite{Model:Agnostic}. On the other hand, model-specific techniques leverage the properties of the underlying model to generate explanations, e.g., \cite{BostroemGurungLindgren}. Explanation methods can produce explanations in the form of plots, such as the Partial Dependence Plot (PDP) and the Accumulated Local Effects (ALE) Plot \citep{ALE_Apley}. Rule-based explanations, e.g., those produced by Anchors \citep{anchors}, are another easy to interpret explanation form. Finally, additive feature importance scores, such as those generated by LIME (Local Interpretable Model-agnostic Explanations) \citep{lime} and SHAP (SHapley Additive exPlanations) \citep{shap}, are another form of explanation.

However, the deployment of explanation methods comes with its challenges. One of the most significant challenges is the computational cost, which can be for several reasons. One main reason is that many methods analyze the behavior of a black-box model by running multiple iterations on different input data points, which can be time-consuming \citep{molnar_brief_history}. 
Also, some explanation methods, e.g., LIME \citep{lime}, involve creating multiple local (white-box) surrogate models.
Furthermore, one of the most influential explanation methods, SHAP, involves approximating Shapley values to compute the marginal importance of the features \citep{shap}. Computing Shapley values requires calculating the contributions of each feature in a given prediction, where the cost grows exponentially with the number of features.

In this work, we address the challenge of handling the computational cost associated with explainable machine learning methods. We propose to reduce the cost by approximating the output of any selected explanation method that produces feature importance scores using an algorithm of lower complexity at inference time and provide some validity guarantees on the produced explanations using the conformal prediction framework \citep{vovk2005algorithmic}. The proposed approach will be thoroughly evaluated to demonstrate its effectiveness in approximating the explanation method while reducing the computational cost.

The main contributions of this study are:

\begin{itemize}
    \item an approach for approximating the explanations to a black box predictions in the form of feature importance scores accompanied with validity guarantees from the conformal prediction framework
    
    \item a set of non-conformity measures for the conformal prediction framework
    
    \item a large-scale empirical investigation on 30 publicly available datasets comparing the performance of two different algorithms in approximating the explanation method, as well as the performance using the proposed non-conformity measures
    
\end{itemize}

In the next section, we provide a background on the conformal prediction framework. In \sectionref{sec:related_work}, we briefly review related work. In \sectionref{sec:method}, we describe the proposed method and propose a set of non-conformity measures designed to reduce the computational cost. \sectionref{sec:evaluation} presents and discusses the results of the large-scale empirical investigation. Finally, \sectionref{CR} recapitulates the main conclusions and outlines future work directions.

\section{Background}

Conformal Prediction (CP) has been introduced as an approach for providing guarantees on the prediction error. A user-predefined confidence score bounds the probability of producing wrong predictions \citep{Reg_CP_RF}. CP was initially introduced as a transductive approach that learns a model for each data instance, which is computationally expensive \citep{Gammerman_1998,Saunders_1999}. Consequently, \cite{vovk2005algorithmic} introduced inductive conformal prediction (ICP), where only one model is induced from the provided data, which is then used for making predictions on new data instances. In the remaining part of this section, we briefly describe inductive conformal prediction for regression models, as well as the possible non-conformity functions.

\subsection{Inductive Conformal Prediction for Regression Models}

Let $\mathcal{X}$ be a feature space and $\mathcal{Y}$ be the target variable. Given a dataset $\mathcal{Z} = \{z_1, z_2, ..., z_n\}$, where $z_i = (x_i, y_i)$, $x_i \in \mathcal{X}$, and $y_i \in \mathbb{R}$. Assuming the provided data are independent and identically distributed (i.i.d), the inductive conformal regression consists of the following main steps:

\begin{enumerate}

    \item Split the provided dataset $\mathcal{Z}$ into a proper training subset $\mathcal{Z}_t = \{z_1, z_2, ..., z_m\}$ and a calibration subset $\mathcal{Z}_c = \{z_{m+1}, z_{m+2}, ..., z_n\}$.
    
    \item The underlying model $h$ is trained using $\mathcal{Z}_t$.
    
    \item For each example $z_i \in \mathcal{Z}_c$, use the non-conformity function to calculate the non-conformity score $\alpha_i$ to get the sequence $\mathcal{S} = \{\alpha_{m+1}, \alpha_{m+2}, . . . , \alpha_{n}\}$. The non-conformity function can be simply the absolute error \citep{Papadopoulos_2002}:

    \begin{equation}
        \alpha_{i} = |y_i - \tilde{y}_i|
    \end{equation}

    where $\tilde{y}_i$ is the predicted outcome by the underlying model $h$.

    \item Given a predefined significance level $\epsilon$ and the sequence of the non-conformity scores $\mathcal{S}$, the smallest $\alpha_{\epsilon} \in \mathcal{S}$ such that:

    \begin{equation}
        \frac{|\{z_i \in \mathcal{Z}_c | \alpha_i < \alpha_{\epsilon}\}| + 1}{|\mathcal{Z}_c| + 1} \geq 1 - \epsilon
    \end{equation}

    Consequently, with a probability of 1-$\epsilon$, the non-conformity score of a new data instance $x_{n+1}$ will be less than or equal to $\alpha_{\epsilon}$, since $\alpha_{\epsilon}$ provides a probabilistic bound for the non-conformity scores at the significance level $\epsilon$. Finally, an interval covering the true prediction with a probability of 1-$\epsilon$ is produced as follows:

    \begin{equation}
         \tilde{\mathscr{Y}}_{n+1} = [\tilde{y}_{n+1} - \alpha_{\epsilon}, \tilde{y}_{n+1} + \alpha_{\epsilon}]
    \end{equation}

\end{enumerate}

\subsection{Normalized Non-Conformity Functions} \label{normalized}

The previous description of the inductive conformal regression results in a fixed interval size for all predictions. However, some predictions are expected to be more accurate than others, and a natural improvement is to predict smaller intervals for more accurate (easy) cases. Therefore, the non-conformity score $\alpha_i$ of the instance $x_i$ can be normalized by a difficulty estimate $\sigma_i$:

    \begin{equation}
            \alpha_{i} = \frac{|y_i - \tilde{y}_i|}{\sigma_i}
    \end{equation}

\noindent For a prediction on a new data instance $x_{n+1}$, the predicted interval is given by:

    \begin{equation}
         \tilde{\mathscr{Y}}_{n+1} = [\tilde{y}_{n+1} - \alpha_{\epsilon}\sigma_{n+1}, \tilde{y}_{n+1} + \alpha_{\epsilon}\sigma_{n+1}]
    \end{equation}

\cite{Papadopoulos_2011} proposed to estimate the difficulty using the $k$-nearest neighbours (KNN). According to \cite{Papadopoulos_2011}, this can be done by computing the sum of the distances between $x_i$ and its $k$-nearest neighbours:

\begin{equation}
     d^k_i = \sum_{\eta \in \mathcal{N}} distance(x_i, x_{\eta})
\end{equation}

\noindent where $\mathcal{N}$ is the set of the $k$-nearest neighbours. Then $d^k_i$ is normalized by the median value of the distances in the training data:

\begin{equation}
    \lambda^k_i = \frac{d^k_i}{median(d_j: z_j \in \mathcal{Z}_t)}
\end{equation}

\noindent Finally, the non-conformity score is computed as follows:

\begin{equation}
    \alpha_{i} = \frac{|y_i - \tilde{y}_i|}{\gamma + \lambda^k_i},
\end{equation}

and

\begin{equation}
    \alpha_{i} = \frac{|y_i - \tilde{y}_i|}{exp(\gamma \lambda^k_i)}.
\end{equation}

\noindent where $\gamma \geq 0$ is a parameter that controls the measure's sensitivity to any changes in $\lambda^k_i$.

Another difficulty estimate proposed by \cite{Papadopoulos_2011} is based on the difference between the labels of the $k$-nearest neighbours measured in their standard deviation. The high agreement among the $k$-nearest neighbours typically means a more accurate prediction. The standard deviation of the labels of the $k$-nearest neighbours is computed as follows:

\begin{equation}
    s^k_i = \sqrt{\frac{1}{k} \sum_{j=1}^{k} (y_{i_j} - \overline{y_{i_{1, 2, ..., k}}})^2},
\end{equation}

\noindent where
\begin{equation}
    \overline{y_{i_{1, 2, ..., k}}} = \frac{1}{k} \sum_{j=1}^{k} y_{i_j}
\end{equation}

\noindent Then $s^k_i$ can also be normalized, similar to what has been proposed with $d^k_i$:

\begin{equation}
    \xi^k_i = \frac{s^k_i}{median(s_j: z_j \in \mathcal{Z}_t)}
\end{equation}

\noindent Consequently, the non-conformity measure can be computed as follows:

\begin{equation}
    \alpha_{i} = \frac{|y_i - \tilde{y}_i|}{\gamma + \xi^k_i},
\end{equation}

\noindent and

\begin{equation}
    \alpha_{i} = \frac{|y_i - \tilde{y}_i|}{exp(\gamma \xi^k_i)}.
\end{equation}

\noindent Finally, $\lambda^k_i$ and $\xi^k_i$ can be combined to define the following non-conformity measures:

\begin{equation}
    \alpha_{i} = \frac{|y_i - \tilde{y}_i|}{\gamma + \lambda^k_i + \xi^k_i},
\end{equation}

\noindent and

\begin{equation}
    \alpha_{i} = \frac{|y_i - \tilde{y}_i|}{exp(\gamma \lambda^k_i) + exp(\rho \xi^k_i)},
\end{equation}

\noindent where $\rho$ controls the sensitivity of the measure to the changes in $\xi^k_i$, similar to $\gamma$ with $\lambda^k_i$

\section{Related Work}
\label{sec:related_work}

In this section, we start by describing two popular explanation methods and clarify why they are computationally expensive. We also provide some pointers to the contributions to providing computationally more efficient machine learning explanation methods. 


Explainable machine learning have recently caught significant attention as a research field, especially since the introduction of the LIME technique for generating local explanations (for a specific prediction) in 2016 \citep{lime}. LIME fits a white-box model on the predictions of the underlying black box on perturbed data points, which are weighted by proximity to the being explained data point. The feature weights of the white box are used to generate an explanation for the prediction. Since LIME provides local explanations, its complexity does not depend on the size of the dataset but on the number of perturbed data instances, the complexity of the underlying black box, the complexity of the used white box, and the length of the produced explanation (number of features to explain a prediction). SHAP \citep{shap} is another prominent local explanation method that assigns importance scores to the features by approximating the Shapley values. Since the exact Shapely values computation requires all possible coalitions of the feature values, SHAP (i.e., Kernel SHAP) approximates the exact Shapley values using sampled feature coalitions, and a linear model is fitted to approximate the Shapley values.

\cite{tree_shap} introduced a faster variant of SHAP for tree-based models, e.g., random forests and gradient-boosted trees. The variant is TreeSHAP, a model-specific alternative to the model-agnostic Kernel SHAP. The time complexity of SHAP can be reduced from $O(T L2^M)$ to $O(TLD^2)$ by using TreeSHAP for a tree-based model \citep{tree_shap}, where $T$ is the number of trees, $L$ is the maximum number of leaves in any tree, $D$ is the maximum depth of any tree, and $M$ is the number of features. \cite{situ-etal} proposed that any off-the-shelf explainer can be distilled into an explainer neural network (L2E, Learning to Explain). L2E focused mainly on imitating the explanations obtained on text classification tasks. Approximating the outcome of the explanation method using a neural network can reduce the time complexity to the level of the neural network at the inference time and can also help with producing stable explanations. However, at the inference time, there are no guarantees regarding the validity of the predicted explanation at the individual word level. FastSHAP was proposed by \cite{jethani2022fastshap} to improve the run-time of the Shapley values approximation-based explanation methods. FastSHAP avoids conducting an optimization process for each data point by learning a parametric function to approximate the Shapley value explanations. For image classification explanations, h-Shap (Hierarchical Shap) \citep{teneggi2022fast} has been proposed as a fast and exact implementation of Shapley coefficients.

\section{Method} \label{sec:method}

This section first describes a method to approximate any explanation method that produces additive feature importance scores. Afterwards, we discuss how to provide validity guarantees using the conformal prediction framework. Finally, we propose three possible non-conformity measures for the conformal regression. 

\subsection{Explanation Method Approximation}

The proposed method can be applied to any explanation method as long as the produced explanations take the form of additive feature importance scores. The explanation method ($\mathcal{A}$) is considered a function ($\mathcal{A}: f(\textbf{x}; t; \Theta) = \textbf{y}$) that can be approximated using a machine learning model ($\mathcal{\tilde{A}}$), where $\textbf{x} \in \mathbb{R}^n$ is the data point with $n$ features, $t$ is the predicted outcome by the black box model, $\Theta$ is some learnable parameters, and $\textbf{y} \in \mathbb{R}^n$ is the vector containing the importance scores of the $n$ features. The approximation model ($\mathcal{\tilde{A}}$) learns a mapping from $(\textbf{x}; t)$ to $\textbf{y}$. Since the target $\textbf{y}$ is a vector with an importance score per feature, the problem can be formulated as a regression problem. There are two possible approaches to solving this regression problem: i) to handle it as a multi-target regression problem,  using, for instance, a neural network, and ii) to handle it as a set of single-target regression problems.  

A development dataset ($\mathcal{X}^{dev}$) is provided for the regression model, where the underlying black box ($\mathcal{B}$) produces predictions $\textbf{t}^{dev}$ for the data points in $\mathcal{X}^{dev}$, and the explanation method ($\mathcal{A}$) generates explanations ($\mathcal{Y}^{dev}$) for the predicted outcomes. For each data point in the development set, a feature vector ($\textbf{x}$) is augmented with its predicted outcome ($t$) to build the augmented development features ($\textbf{x}^{'} = \textbf{x} \cup t$). The augmented development set $\mathcal{X}^{'dev}$ altogether with produced explanations $\mathcal{Y}^{dev}$ form $\mathcal{Z}^{dev} = \{(\textbf{x}^{'}_{1}, \textbf{y}_1), (\textbf{x}^{'}_{2}, \textbf{y}_2), ..., (\textbf{x}^{'}_{n}, \textbf{y}_n)\}$. $\mathcal{\tilde{A}}$ is learned by fitting the regression model on $\mathcal{Z}^{dev}$.



\subsection{Validity Guarantees}

The learned regression model $\mathcal{\tilde{A}}$ predicts an importance score per feature, and we can simply control the error level of each feature independently from the others using the conformal prediction framework. In other words, each predicted importance score is considered a separate regression problem, and the conformal regression will be applied to control the error level. Alternatively, the problem can be handled as a conformal multi-target regression similar to the method proposed by \cite{messoudi20a}. However, we will leave the alternative approach for future work. Consequently, a calibration dataset $\mathcal{X}^{cal}$ is provided and augmented with the class label acquired from $\mathcal{B}$ to obtain $\mathcal{X}^{'cal}$, which is provided to $\mathcal{\tilde{A}}$ to generate importance scores for all the data points in the calibration set $\mathcal{X}^{cal}$ (predict $\mathcal{\tilde{Y}}^{cal}$). Using the ground truth $\mathcal{Y}^{cal}$ obtained from $\mathcal{A}$, a non-conformity score $\alpha_{j}^{f}$ is computed for each feature $f$ for each example $\textbf{x}_j$ in $\mathcal{X}^{cal}$. Let $\alpha_{\epsilon}^{f}$ be the score of feature $f$ at a significance level $\epsilon$. At prediction time, all values with a distance greater than $\alpha_{\epsilon}^{f}$ are excluded: $\tilde{\mathscr{Y}}_j^f = [\tilde{y}_j^{f}-\alpha_{\epsilon}^{f}, \tilde{y}_j^{f}+\alpha_{\epsilon}^{f}]$.

\subsection{Non-Conformity Measures} \label{non_conf}

As described in Section~\ref{normalized}, the non-conformity score can be normalized using a difficulty estimate $\sigma_j$, which can be computed using, for example, a trained model or the distance to the $k$-nearest neighbors. However, such difficulty estimates are computed at both the calibration and the inference time per instance and can be computationally expensive. Therefore we propose the following difficulty estimation functions:

\begin{enumerate}

    \item \textbf{Minimum distance to the distributions:} we assume there is a mixture of distributions where each class $\mathcal{C}$ in the training data represents one distribution. Consequently, we fit a Gaussian mixture model \citep{NIPS1999_97d98119} on the training data. Then we use the mean $\mu_{\mathcal{C}}$ and the covariance $\Sigma_{\mathcal{C}}$ of each distribution to compute the Mahalanobis distance \citep{mahalanobis1936} between a data point and each distribution:

    \begin{equation}
        d_{j \mathcal{C}} = \sqrt{(x_j - \mu_{\mathcal{C}})^T \Sigma_{\mathcal{C}}^{-1} (x_j - \mu_{\mathcal{C}})}
    \end{equation}

    Then the minimum distance is used as a difficulty estimate:

    \begin{equation}
        \sigma_j = \log (\operatorname*{arg\,min}_{\mathcal{C}}(d_{j \mathcal{C}}) +1)
    \end{equation}

    \item \textbf{Average distance to the distributions:} we also assume a mixture of class distributions in the training data, but here we take the average of all distances from the data point to the distributions:

    \begin{equation}
        \sigma_j = \log (\frac{1}{n}\mathlarger{\sum_{\mathcal{C}=1}^{n} d_{j \mathcal{C}}} +1)
    \end{equation}

    \item \textbf{The prediction confidence:} we assume that if the prediction towards one class has high algorithmic confidence, then it is an easy example for the black box model. Therefore, we propose the following difficulty estimate:

    \begin{equation}
        \sigma_j = 1 - max(\mathcal{P}_{\mathcal{C}})
    \end{equation}

    where $\mathcal{P}_{\mathcal{C}}$ is the predicted score of class $\mathcal{C}$, and for the binary classification we use the following:

    \begin{equation}
        \sigma_j = 1 - | \mathcal{P} - 0.5 |
    \end{equation}

    where $\mathcal{P}$ is the predicted score.
\end{enumerate}

\section{Empirical Evaluation} \label{sec:evaluation}

This section compares the generated explanations using two distinct regression models. The quality of the generated explanations is assessed with respect to the execution time as well as the produced interval sizes using the conformal prediction framework, where a tighter interval size implies a better approximation. We conduct two sets of experiments. We also evaluate the explanations using the proposed non-conformity measures. 

\subsection{Experimental Setup}

In the following experiments, the proposed explanation algorithms are evaluated using 30 publicly available datasets\footnote{All the datasets were obtained from \url{https://www.openml.org}}. Each dataset is split into a training set, calibration set, and test set. The calibration set is used to compute the non-conformity scores and find the confidence percentile score, the training split is used to train the black-box model, and the test split is used to evaluate the generated explanations. The conformal regressors were generated using the \textsf{crepes}\footnote{\url{https://github.com/henrikbostrom/crepes}} Python package \citep{crepes}. Two algorithms are used for the explanation technique approximation: XGBoost and multi-layer perceptron (MLP). In the XGBoost experiments, one regressor is trained per feature, while in the case of MLP, one regression model is trained to predict all the feature importance scores. The XGBoost model employs a learning rate of 0.1, 600 estimators, and 0.01 for the regularization parameter. The MLP model has two layers, each of 1024 units and a Relu activation function. The MLP is trained with early stopping and 0.1 validation fraction.

The underlying black-box model is learned through an XGBoost algorithm. The hyperparameters of XGBoost are tuned through grid search. The hyperparameters include the learning rate, the number of estimators, and the regularization parameter. The categorical features are binarized in the data preprocessing phase using one-hot encoding. The model is trained with each combination of hyperparameters on the training set and evaluated on the calibration set. The XGBoost model is trained using the best-performing set of hyperparameters. TreeSHAP is used as an explanation technique to the underlying black-box model.

\subsection{Experimental Results}

\subsubsection{Execution Time}

The execution time is measured on the test set of each dataset and recorded in seconds. All experiments have been performed in a Python environment on an Intel(R) Core(TM) i9-10885H CPU @ 2.40GHz system. The baseline for comparing execution times is set by TreeSHAP, a faster variant of SHAP created for tree-based models. In this experiment we compare TreeSHAP to MLP and XGBoost regressors with the proposed non-conformity measures in subsection \ref{non_conf}: the minimum distance to the distributions (Min. Dist.), the average distance to the distributions (Avg. Dist.), the prediction confidence (Pred. Conf.), the $k$-nearest neighbours (KNN), and without any difficulty estimate (Baseline MLP and Baseline XGBoost). 

\begin{figure*}
    \centering
    \includegraphics[width=1.\textwidth]{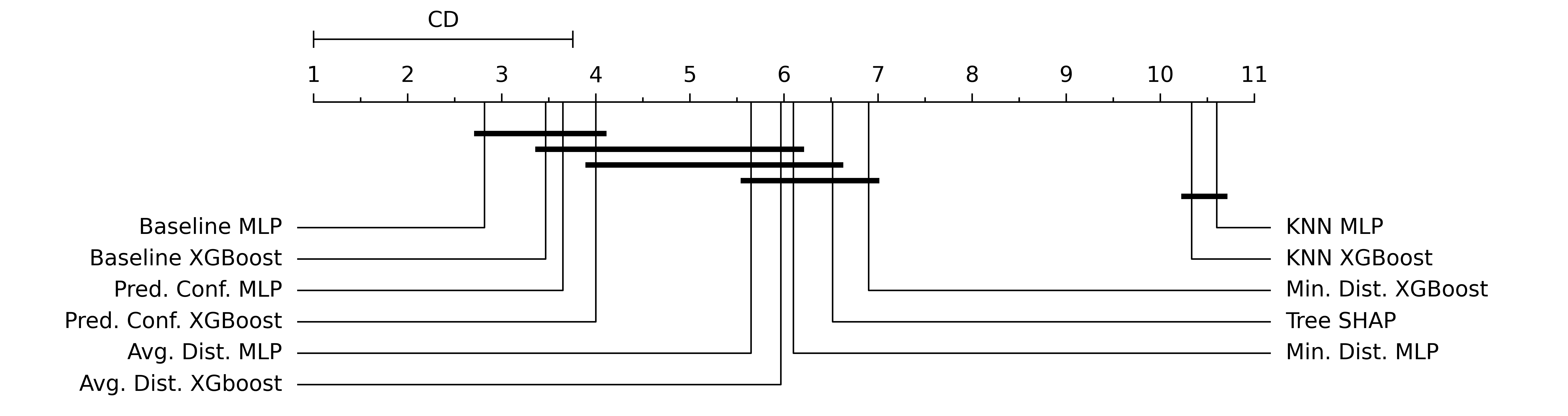}
    \caption{The average rank of the compared regression models and TreeSHAP on the 30 datasets with respect to the execution time (in seconds), where the critical difference (CD) represents the largest difference that is not statistically significant.}
    \label{fig:exec_time}
\end{figure*}

The Friedman test \citep{Friedman_test} is applied to assess the null hypothesis that there is no significant difference in the time needed to generate explanations using TreeSHAP explainer or any of the XGBoost and MLP regression models with the different non-conformity measures. The result of the Friedman test allows for rejecting the null hypothesis at the 0.05 leve, thereby confirming a significant difference in the execution time. Subsequently, the post-hoc Nemenyi test \citep{nemenyi:distribution-free} is employed to determine significant pairwise differences, which are presented in Figure \ref{fig:exec_time}. The results show that XGBoost and MLP without a difficulty estimate and MLP with Pred. Conf.) significantly outperform TreeSHAP. However, no significant difference in execution time has been observed between TreeSHAP and MLP (with Min. Dist. or Avg. Dist.) or XGBoost (with Min. Dist., Avg. Dist., or Pred. Conf.). On the other hand, MLP and XGBoost with KNN are significantly outperformed by TreeSHAP. Furthermore, it has been observed that the execution time of XGBoost is affected by the number of features since we train a regressor per feature, while MLP is affected by the number of instances in a dataset. The detailed results are shown in \tableref{table:exec_time}.

\begin{table*}
\centering
\begin{adjustbox}{angle=90, width=1.0\textwidth}
\small
\begin{threeparttable}
\caption{The time needed to generate explanations for all data points in the test set. The execution time is measured in seconds. The compared methods are TreeSHAP and approximated explanations using XGBoost and MLP with different non-conformity measures. The best-performing model is \colorbox[HTML]{A8B6E0}{colored in blue}, and the second best-performing is \colorbox[HTML]{E0EAF8}{colored in light blue}.
}
\begin{tabular}{l c c c | c c c c c | c c c c c}
    \toprule
    \rowcolor[HTML]{EFEFEF} 
    \multicolumn{1}{c}{\cellcolor[HTML]{EFEFEF}} & \multicolumn{1}{c}{\cellcolor[HTML]{EFEFEF}}  & \multicolumn{1}{c}{\cellcolor[HTML]{EFEFEF}}    & \multicolumn{1}{c}{\cellcolor[HTML]{EFEFEF}}  & \multicolumn{5}{c}{\cellcolor[HTML]{EFEFEF}XGBoost} & \multicolumn{5}{c}{\cellcolor[HTML]{EFEFEF}Multi-Layer Perceptron} \\ \cline{5-14} 
    \rowcolor[HTML]{EFEFEF} 
    \multicolumn{1}{c}{\multirow{-2}{*}{\cellcolor[HTML]{EFEFEF}Dataset}} & \multicolumn{1}{c}{\multirow{-2}{*}{\cellcolor[HTML]{EFEFEF}Test Set Size}} & \multicolumn{1}{c}{\multirow{-2}{*}{\cellcolor[HTML]{EFEFEF}Num. Features}} & \multicolumn{1}{c|}{\multirow{-2}{*}{\cellcolor[HTML]{EFEFEF}TreeSHAP}} & Min. Dist. & Avg. Dist. & KNN & Pred. Conf. & Baseline & Min. Dist. & Avg. Dist. & KNN & Pred. Conf. & Baseline \\
    \cmidrule(lr){1-14}
    Abalone & 835 & 10 & 0.129 & 0.08 & 0.076 & 0.401 & \colorbox[HTML]{E0EAF8}{0.042} & \colorbox[HTML]{A8B6E0}{0.04} & 0.056 & 0.059 & 0.326 & 0.048 & \colorbox[HTML]{E0EAF8}{0.042} \\ \cmidrule(lr){1-14}
    Ada Prior & 912 & 100 & 0.106 & 0.519 & 0.442 & 4.615 & 0.245 & 0.238 & 0.407 & 0.402 & 4.626 & \colorbox[HTML]{A8B6E0}{0.064} & \colorbox[HTML]{A8B6E0}{0.064} \\ \cmidrule(lr){1-14}
    Adult & 2442 & 105 & 0.469 & 1.061 & 1.03 & 16.023 & 0.385 & 0.381 & 1.074 & 1.075 & 20.143 & \colorbox[HTML]{E0EAF8}{0.155} & \colorbox[HTML]{A8B6E0}{0.138} \\ \cmidrule(lr){1-14}
    Bank 32 nh & 1229 & 32 & 0.286 & 0.152 & 0.158 & 1.773 & 0.128 & 0.127 & 0.104 & 0.105 & 2.041 & \colorbox[HTML]{A8B6E0}{0.08} & \colorbox[HTML]{A8B6E0}{0.08} \\ \cmidrule(lr){1-14}
    Breast Cancer & 3937 & 9 & 0.719 & 0.158 & 0.156 & 0.913 & \colorbox[HTML]{E0EAF8}{0.079} & \colorbox[HTML]{A8B6E0}{0.075} & 0.29 & 0.279 & 1.245 & 0.228 & 0.22 \\ \cmidrule(lr){1-14}
    Churn & 1000 & 32 & 0.192 & 0.112 & 0.11 & 1.357 & 0.077 & 0.08 & 0.088 & 0.081 & 1.725 & \colorbox[HTML]{E0EAF8}{0.069} & \colorbox[HTML]{A8B6E0}{0.067} \\ \cmidrule(lr){1-14}
    Credit Card Fraud & 7120 & 30 & \colorbox[HTML]{A8B6E0}{0.195} & 0.574 & 0.428 & 36.168 & \colorbox[HTML]{E0EAF8}{0.315} & 0.319 & 3.384 & 3.333 & 47.278 & 3.379 & 3.339 \\ \cmidrule(lr){1-14}
    Delta Ailerons & 1782 & 5 & 0.407 & 0.08 & 0.069 & 0.218 & \colorbox[HTML]{E0EAF8}{0.033} & \colorbox[HTML]{A8B6E0}{0.027} & 0.096 & 0.097 & 0.23 & 0.078 & 0.078 \\ \cmidrule(lr){1-14}
    Delta Elevators & 1903 & 6 & 0.109 & 0.088 & 0.082 & 0.315 & \colorbox[HTML]{A8B6E0}{0.03} & \colorbox[HTML]{E0EAF8}{0.031} & 0.098 & 0.108 & 0.31 & 0.084 & 0.083 \\ \cmidrule(lr){1-14}
    Electricity & 4531 & 14 & 0.899 & 0.242 & 0.268 & 1.818 &  \colorbox[HTML]{E0EAF8}{0.118} & \colorbox[HTML]{A8B6E0}{0.11} & 0.804 & 0.803 & 2.051 & 0.741 & 0.722 \\ \cmidrule(lr){1-14}
    Elevators & 2490 & 18 & 0.878 & 0.151 & 0.136 & 3.015 & \colorbox[HTML]{E0EAF8}{0.095} & \colorbox[HTML]{A8B6E0}{0.093} & 0.135 & 0.138 & 2.805 & 0.109 & 0.1 \\ \cmidrule(lr){1-14}
    Higgs & 4902 & 28 & 2.104 & 0.408 & 0.311 & 16.223 & \colorbox[HTML]{A8B6E0}{0.229} & \colorbox[HTML]{E0EAF8}{0.283} & 1.257 & 1.24 & 16.784 & 1.201 & 1.163 \\ \cmidrule(lr){1-14}
    JM1 & 1088 & 21 & 0.143 & 0.105 & 0.094 & 0.941 & 0.081 & 0.073 & 0.065 & 0.068 & 0.854 & \colorbox[HTML]{E0EAF8}{0.037} & \colorbox[HTML]{A8B6E0}{0.034} \\ \cmidrule(lr){1-14}
    Madelon & 520 & 500 & 0.144 & 1.484 & 1.439 & 16.78 & 1.305 & 1.335 & 0.202 & 0.197 & 14.767 & \colorbox[HTML]{E0EAF8}{0.035} & \colorbox[HTML]{A8B6E0}{0.031} \\ \cmidrule(lr){1-14}
    Magic Telescope & 1902 & 10 & 0.263 & 0.088 & 0.092 & 0.476 & \colorbox[HTML]{E0EAF8}{0.044} & \colorbox[HTML]{A8B6E0}{0.041} & 0.111 & 0.102 & 0.576 & 0.082 & 0.083 \\ \cmidrule(lr){1-14}
    Mozilla4 & 947 & 38 & 0.086 & 0.145 & 0.135 & 1.534 & 0.095 & 0.092 & 0.064 & 0.058 & 1.452 & \colorbox[HTML]{E0EAF8}{0.044} & \colorbox[HTML]{A8B6E0}{0.036} \\ \cmidrule(lr){1-14}
    MC1 & 994 & 38 & \colorbox[HTML]{A8B6E0}{0.023} & 0.111 & 0.102 & 1.605 & 0.089 & 0.09 & 0.065 & 0.064 & 1.489 & 0.052 & \colorbox[HTML]{E0EAF8}{0.042} \\ \cmidrule(lr){1-14}
    Numerai28.6 & 4816 & 21 & 2.551 & 0.368 & 0.274 & 11.788 & \colorbox[HTML]{E0EAF8}{0.198} & \colorbox[HTML]{A8B6E0}{0.151} & 0.528 & 0.531 & 11.827 & 0.463 & 0.453 \\ \cmidrule(lr){1-14}
    PC2 & 1118 & 36 & \colorbox[HTML]{A8B6E0}{0.026} & 0.11 & 0.104 & 1.562 & 0.086 & 0.082 & 0.068 & 0.069 & 1.409 & 0.053 & \colorbox[HTML]{E0EAF8}{0.048} \\ \cmidrule(lr){1-14}
    Phishing & 1105 & 68 & 0.082 & 0.235 & 0.231 & 3.35 & 0.207 & 0.211 & 0.08 & 0.072 & 2.6 & \colorbox[HTML]{A8B6E0}{0.056} & \colorbox[HTML]{A8B6E0}{0.056} \\ \cmidrule(lr){1-14}
    Phonemes & 811 & 5 & 0.123 & 0.031 & 0.03 & 0.127 & \colorbox[HTML]{E0EAF8}{0.023} & \colorbox[HTML]{A8B6E0}{0.017} & 0.053 & 0.049 & 0.126 & 0.038 & 0.036 \\ \cmidrule(lr){1-14}
    Pollen & 770 & 5 & 0.047 & 0.032 & 0.022 & 0.116 & \colorbox[HTML]{E0EAF8}{0.015} & \colorbox[HTML]{A8B6E0}{0.013} & 0.057 & 0.056 & 0.159 & 0.047 & 0.04 \\ \cmidrule(lr){1-14}
    Satellite & 1147 & 36 & \colorbox[HTML]{A8B6E0}{0.049} & 0.115 & 0.118 & 1.814 & 0.082 & 0.081 & 0.09 & 0.082 & 1.855 & 0.075 & \colorbox[HTML]{E0EAF8}{0.072} \\ \cmidrule(lr){1-14}
    Scene & 602 & 304 & \colorbox[HTML]{A8B6E0}{0.048} & 0.825 & 0.801 & 9.984 & 0.731 & 0.699 & 0.361 & 0.339 & 12.919 & 0.058 & \colorbox[HTML]{E0EAF8}{0.056} \\ \cmidrule(lr){1-14}
    Spambase & 690 & 57 & 0.16 & 0.162 & 0.14 & 1.707 & 0.144 & 0.126 & 0.074 & 0.07 & 2.751 & \colorbox[HTML]{E0EAF8}{0.056} & \colorbox[HTML]{A8B6E0}{0.054} \\ \cmidrule(lr){1-14}
    Speed Dating & 1257 & 500 & 0.231 & 2.285 & 2.335 & 39.266 & 2.047 & 2.018 & 0.945 & 0.851 & 46.444 & \colorbox[HTML]{E0EAF8}{0.116} & \colorbox[HTML]{A8B6E0}{0.114} \\ \cmidrule(lr){1-14}
    Telco Customer Churn & 1056 & 45 & 0.386 & 0.117 & 0.121 & 1.891 & 0.102 & 0.1 & 0.088 & 0.082 & 2.501 & \colorbox[HTML]{E0EAF8}{0.064} & \colorbox[HTML]{A8B6E0}{0.062} \\ \cmidrule(lr){1-14}
    Tic Tac Toe & 4921 & 27 & 1.126 & 0.385 & 0.347 & 14.936 & \colorbox[HTML]{E0EAF8}{0.259} & \colorbox[HTML]{A8B6E0}{0.198} & 0.363 & 0.352 & 20.077 & 0.29 & 0.289 \\ \cmidrule(lr){1-14}
    Vehicle sensIT & 9853 & 100 & 5.994 & 4.558 & 4.098 & 262.276 & \colorbox[HTML]{A8B6E0}{1.729} & \colorbox[HTML]{E0EAF8}{1.78} & 9.137 & 9.03 & 323.104 & 5.767 & 5.735 \\ \cmidrule(lr){1-14}
    Waveform-5000 & 1000 & 40 & 0.194 & 0.542 & 0.172 & 3.286 & 0.159 & 0.156 & 0.096 & 0.084 & 2.503 & \colorbox[HTML]{E0EAF8}{0.07} & \colorbox[HTML]{A8B6E0}{0.068} \\ \cmidrule(lr){1-14}
    \rowcolor[HTML]{EFEFEF}
    Average rank & -- & -- & 6.517 & 6.9 & 5.967 & 10.33 & 4 & \colorbox[HTML]{E0EAF8}{3.467} & 6.1 & 5.65 & 10.6 & 3.65 & \colorbox[HTML]{A8B6E0}{2.82} \\ \bottomrule

\end{tabular}

\label{table:exec_time}
\end{threeparttable}
\end{adjustbox}
\end{table*}

\subsubsection{Interval Size}

Since the conformal regression provides the needed validity guarantees and the correct feature importance scores are ensured to be covered by the produced intervals with a specified confidence level $c$, then the size of the generated intervals becomes the metric of how useful the predictions are, where tighter intervals are more informative. The scale of the importance scores generated differs between datasets. Therefore, the interval sizes displayed in \tableref{table:experiments}, as well as Tables \ref{table:experiments_top10} and \ref{table:experiments_top5} in \appendixref{appendix}, are normalized by the difference between each feature's maximum and minimum importance values, as generated by the TreeSHAP explainer.

We display the results averaged across all importance scores within each dataset in \tableref{table:experiments}, where the models are obtained through XGBoost and MLP using the proposed non-conformity measures in subsection \ref{non_conf}.

\begin{figure*}[htbp]
    \centering
    \includegraphics[width=1.\textwidth]{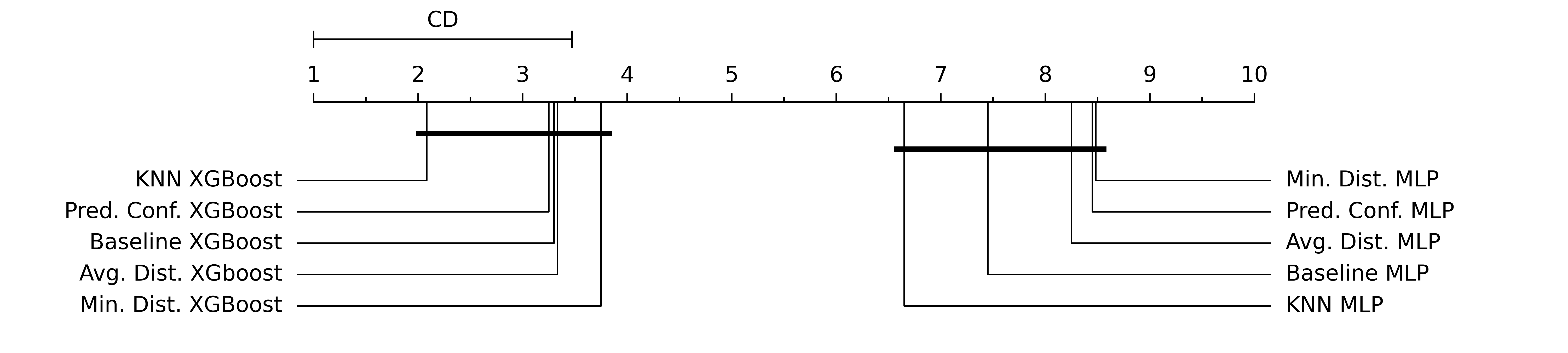}
    \caption{The average rank of the compared regression models on the 30 datasets with respect to the interval size of \textbf{all features} (a lower rank is better), where the critical difference (CD) represents the largest difference that is not statistically significant.}
    \label{fig:ranks_all}
\end{figure*}

The Friedman test is applied to test the null hypothesis that there is no significant difference in the resulting interval sizes produced using XGBoost and MLP with the different non-conformity measures. The Friedman test rejects the null hypothesis at the 0.05 leve, meaning there is a significant difference in the produced interval sizes. The post-hoc Nemenyi test is applied to determine the significant pairwise differences, which are summarized in Figure \ref{fig:ranks_all}. The results show a significant difference between XGBoost and MLP. However, no significant difference in interval sizes is observed between the non-conformity measures applied to any of the two algorithms. 

Since, in many cases, the importance scores are not evenly distributed over features, and the importance scores of a few top features are remarkably higher than the remaining ones, it can be sufficient for the user to use the top features in order to explain the prediction. Therefore, we also display the detailed results using the top 10 and 5 features. The importance scores of each feature are averaged over the test set, and the top 10 and top 5 features are selected using the average values. The average size of the intervals of the top 10 and 5 features are reported in \tableref{table:experiments_top10} and \tableref{table:experiments_top5} in \appendixref{appendix}, respectively. Friedman's and Nemenyi's pairwise significance tests are also applied, and the results are similar to those reported using \tableref{table:experiments}. The pairwise tests are summarized in Figure \ref{fig:ranks_top10} and Figure \ref{fig:ranks_top5} in \appendixref{appendix}, respectively.

\begin{figure}[htbp]
  \centering
  \begin{minipage}[t]{0.48\linewidth}
    \centering
    \includegraphics[width=\linewidth]{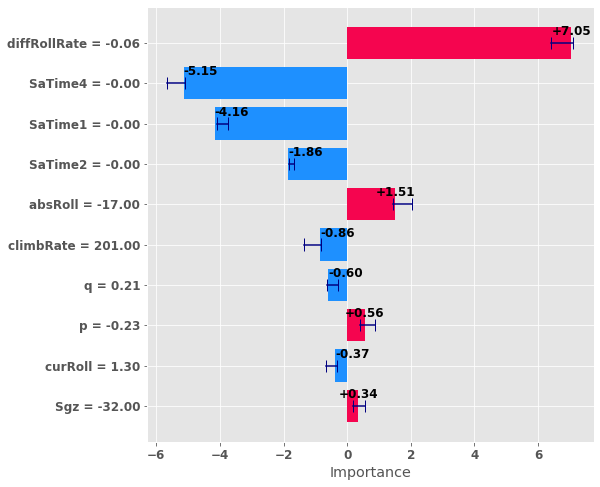}
    \caption{XGBoost with Prob. Conf.}
    \vspace{0.5cm}
    \includegraphics[width=\linewidth]{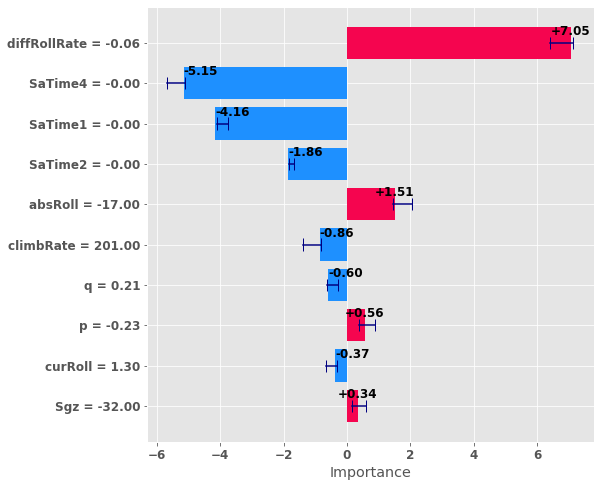}
    \caption{XGBoost without difficulty estimate}

  \end{minipage}
  \hspace{0.2cm}
  \begin{minipage}[t]{0.48\linewidth}
    \centering
    \includegraphics[width=\linewidth]{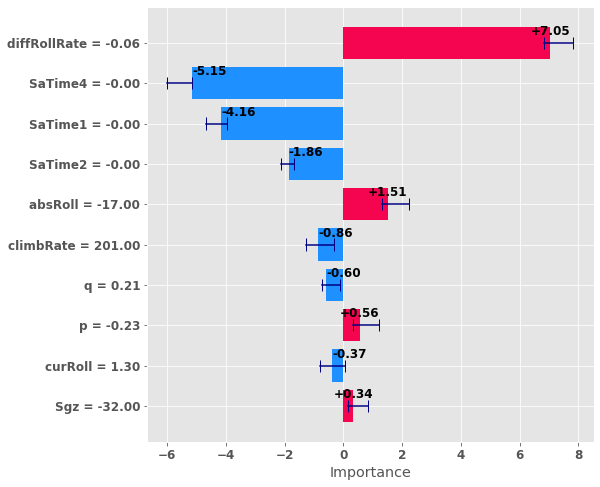}
    \caption{MLP with Prob. Conf.}
    \vspace{0.5cm}
    \includegraphics[width=\linewidth]{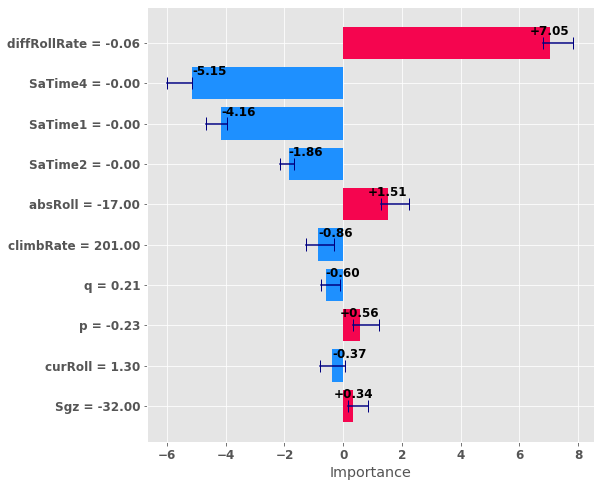}
    \caption{MLP without difficulty estimate}
    \label{fig:mlp_baseline}

  \end{minipage}
\end{figure}

The figures from \ref{fig:xgb_min} to \ref{fig:mlp_baseline} illustrate the predicted intervals using both XGBoost and MLP with the proposed difficulty estimates on one data instance from the Elevators dataset. The bars in the figures are the importance scores generated by the underlying explainer, and the predicted intervals are added to each importance score bar. We plot only the top 10 important features for ease of presentation.

\section{Concluding Remarks} \label{CR}

We have proposed a method to approximate any explanation technique that produces explanations in the form of additive feature importance scores in order to reduce the computational cost of such techniques. We also applied the conformal prediction framework to provide validity guarantees on the approximated explanations. Moreover, a set of non-conformity measures is also proposed to keep the computational cost needed to estimate the sample difficulty low using, for example, KNN. We have presented results from a large-scale empirical investigation comparing two different algorithms (XGBoost and multi-layer perceptron) using the proposed non-conformity measures. The performance of proposed non-conformity measures is compared with the results when KNN is used and when no difficulty estimate is applied. The results show no significant difference between any of the compared non-conformity measures with respect to the interval size. However, the proposed difficulty estimates significantly outperform the ones using KNN with respect to the execution time. The results also show that using an XGBoost regressor per feature can produce a more accurate approximation than using one multi-target MLP regressor for all features. 

A possible direction for future work is to apply the conformal prediction framework to provide validity guarantees for entire explanations instead of providing validity guarantees per feature. Accordingly, an approach similar to conformal multi-target regression \citep{messoudi20a} can be applied.

\begin{table*}
\centering
\begin{adjustbox}{angle=0, width=1.0\textwidth}
\small
\begin{threeparttable}
\caption{The average confidence interval sizes using \textbf{all features}. The generated intervals cover the true importance scores, as predicted by the underlying explainer, with 0.95 confidence. The compared algorithms are XGBoost and MLP, with different non-conformity measures. The best-performing model is \colorbox[HTML]{A8B6E0}{colored in blue}, and the second best-performing is \colorbox[HTML]{E0EAF8}{colored in light blue}.
}
\begin{tabular}{l c c c c c | c c c c c}
    \toprule
    \rowcolor[HTML]{EFEFEF} 
    \multicolumn{1}{c}{\cellcolor[HTML]{EFEFEF}}                          & \multicolumn{5}{c}{\cellcolor[HTML]{EFEFEF}XGBoost} & \multicolumn{5}{c}{\cellcolor[HTML]{EFEFEF}Multi-Layer Perceptron} \\ \cline{2-11} 
    \rowcolor[HTML]{EFEFEF} 
    \multicolumn{1}{c}{\multirow{-2}{*}{\cellcolor[HTML]{EFEFEF}Dataset}} & Min. Dist. & Avg. Dist. & KNN & Pred. Conf. & Baseline & Min. Dist. & Avg. Dist. & KNN & Pred. Conf. & Baseline \\
    \cmidrule(lr){1-11}
    Abalone & 0.219 & 0.116 & \colorbox[HTML]{A8B6E0}{0.075} & \colorbox[HTML]{E0EAF8}{0.089} & 0.094 & 0.528 & 0.304 & 0.174 & 0.223 & 0.235 \\ \cmidrule(lr){1-11}
    Ada Prior & \colorbox[HTML]{E0EAF8}{0.027} & \colorbox[HTML]{E0EAF8}{0.027} & \colorbox[HTML]{A8B6E0}{0.023} & 0.028 & \colorbox[HTML]{E0EAF8}{0.027} & 12.766 & 12.593 & 12.384 & 13.845 & 12.543 \\ \cmidrule(lr){1-11}
    Adult & \colorbox[HTML]{A8B6E0}{0.018} & \colorbox[HTML]{A8B6E0}{0.018} & \colorbox[HTML]{A8B6E0}{0.018} & 0.02 & \colorbox[HTML]{E0EAF8}{0.019} & 3.886 & 3.824 & 3.613 & 4.459 & 3.8 \\ \cmidrule(lr){1-11}
    Bank 32 nh & \colorbox[HTML]{A8B6E0}{0.191} & \colorbox[HTML]{E0EAF8}{0.192} & 0.216 & 0.194 & 0.193 & 0.32 & 0.321 & 0.364 & 0.333 & 0.319 \\ \cmidrule(lr){1-11}
    Breast Cancer & \colorbox[HTML]{E0EAF8}{0.054} & 0.07 & \colorbox[HTML]{A8B6E0}{0.05} & 0.065 & 0.066 & 0.147 & 0.195 & 0.097 & 0.126 & 0.128 \\ \cmidrule(lr){1-11}
    Churn & 0.102 & \colorbox[HTML]{E0EAF8}{0.1} & \colorbox[HTML]{E0EAF8}{0.1} & \colorbox[HTML]{A8B6E0}{0.096} & \colorbox[HTML]{E0EAF8}{0.1} & 24.668 & 24.683 & 22.931 & 25.098 & 25.484\\ \cmidrule(lr){1-11}
    Credit Card Fraud & \colorbox[HTML]{A8B6E0}{0.053} & \colorbox[HTML]{A8B6E0}{0.053} & \colorbox[HTML]{E0EAF8}{0.057} & \colorbox[HTML]{A8B6E0}{0.053} & \colorbox[HTML]{A8B6E0}{0.053} & 0.306 & 0.303 & 0.234 & 0.207 & 0.207 \\ \cmidrule(lr){1-11}
    Delta Ailerons & 0.153 & 0.067 & \colorbox[HTML]{A8B6E0}{0.054} & \colorbox[HTML]{E0EAF8}{0.061} & 0.066 & 0.175 & 0.144 & 0.137 & 0.133 & 0.141 \\ \cmidrule(lr){1-11}
    Delta Elevators & 0.082 & 0.08 & \colorbox[HTML]{A8B6E0}{0.073} & \colorbox[HTML]{E0EAF8}{0.079} & 0.08 & 0.154 & 0.151 & 0.152 & 0.158 & 0.154 \\ \cmidrule(lr){1-11}
    Electricity & \colorbox[HTML]{E0EAF8}{0.103} & \colorbox[HTML]{E0EAF8}{0.103} & \colorbox[HTML]{A8B6E0}{0.102} & 0.104 & \colorbox[HTML]{E0EAF8}{0.103} & 0.181 & 0.181 & 0.191 & 0.185 & 0.181 \\ \cmidrule(lr){1-11}
    Elevators & \colorbox[HTML]{E0EAF8}{0.064} & \colorbox[HTML]{A8B6E0}{0.063} & \colorbox[HTML]{A8B6E0}{0.063} & \colorbox[HTML]{E0EAF8}{0.064} & \colorbox[HTML]{A8B6E0}{0.063} & 6.109 & 6.107 & 5.043 & 5.776 & 5.346 \\ \cmidrule(lr){1-11}
    Higgs & \colorbox[HTML]{A8B6E0}{0.084} & \colorbox[HTML]{A8B6E0}{0.084} & 0.092 & 0.09 & \colorbox[HTML]{E0EAF8}{0.086} & 0.133 & 0.133 & 0.143 & 0.141 & 0.132 \\ \cmidrule(lr){1-11}
    JM1 & 0.119 & 0.12 & \colorbox[HTML]{A8B6E0}{0.099} & \colorbox[HTML]{E0EAF8}{0.115} & 0.12 & 0.39 & 0.256 & 0.198 & 0.215 & 0.218 \\ \cmidrule(lr){1-11}
    Madelon & \colorbox[HTML]{E0EAF8}{0.128} & \colorbox[HTML]{E0EAF8}{0.128} & \colorbox[HTML]{A8B6E0}{0.127} & \colorbox[HTML]{A8B6E0}{0.127} & \colorbox[HTML]{E0EAF8}{0.128} & 2.775 & 2.77 & 2.752 & 2.967 & 2.792 \\ \cmidrule(lr){1-11}
    Magic Telescope & \colorbox[HTML]{A8B6E0}{0.1} & \colorbox[HTML]{A8B6E0}{0.1} & 0.107 & 0.105 & \colorbox[HTML]{E0EAF8}{0.101} & 0.271 & 0.209 & 0.195 & 0.182 & 0.172 \\ \cmidrule(lr){1-11}
    Mozilla4 & 0.026 & \colorbox[HTML]{E0EAF8}{0.024} & \colorbox[HTML]{A8B6E0}{0.015} & \colorbox[HTML]{E0EAF8}{0.024} & \colorbox[HTML]{E0EAF8}{0.024} & 0.995 & 0.998 & 0.961 & 1.155 & 1.158 \\ \cmidrule(lr){1-11}
    MC1 & \colorbox[HTML]{E0EAF8}{0.024} & 0.026 & \colorbox[HTML]{A8B6E0}{0.02} & 0.025 & 0.026 & 4.837 & 4.998 & 3.746 & 5.034 & 5.116 \\ \cmidrule(lr){1-11}
    Numerai28.6 & \colorbox[HTML]{E0EAF8}{0.065} & \colorbox[HTML]{E0EAF8}{0.065} & \colorbox[HTML]{A8B6E0}{0.064} & 0.067 & 0.067 & 0.121 & 0.122 & 0.116 & 0.123 & 0.122 \\ \cmidrule(lr){1-11}
    PC2 & \colorbox[HTML]{E0EAF8}{0.012} & \colorbox[HTML]{E0EAF8}{0.012} & \colorbox[HTML]{A8B6E0}{0.01} & \colorbox[HTML]{E0EAF8}{0.012} & \colorbox[HTML]{E0EAF8}{0.012} & 7.168 & 6.066 & 4.376 & 5.411 & 5.475 \\ \cmidrule(lr){1-11}
    Phishing & 0.03 & \colorbox[HTML]{E0EAF8}{0.029} & \colorbox[HTML]{A8B6E0}{0.025} & 0.03 & \colorbox[HTML]{E0EAF8}{0.029} & 20.439 & 20.449 & 20.024 & 20.969 & 20.484 \\ \cmidrule(lr){1-11}
    Phonemes & 0.176 & 0.172 & 0.182 & \colorbox[HTML]{A8B6E0}{0.16} & \colorbox[HTML]{E0EAF8}{0.166} & 0.296 & 0.258 & 0.281 & 0.252 & 0.245 \\ \cmidrule(lr){1-11}
    Pollen & \colorbox[HTML]{E0EAF8}{0.11} & 0.111 & \colorbox[HTML]{A8B6E0}{0.096} & 0.116 & 0.114 & 0.198 & 0.2 & 0.168 & 0.221 & 0.212 \\ \cmidrule(lr){1-11}
    Satellite & 0.031 & 0.031 & \colorbox[HTML]{A8B6E0}{0.022} & \colorbox[HTML]{E0EAF8}{0.03} & \colorbox[HTML]{E0EAF8}{0.03} & 2.655 & 2.265 & 1.404 & 1.763 & 1.777 \\ \cmidrule(lr){1-11}
    Scene & 0.096 & 0.095 & \colorbox[HTML]{A8B6E0}{0.087} & \colorbox[HTML]{E0EAF8}{0.091} & \colorbox[HTML]{E0EAF8}{0.091} & 10.161 & 9.831 & 9.224 & 9.765 & 9.708 \\ \cmidrule(lr){1-11}
    Spambase & 0.065 & 0.064 & \colorbox[HTML]{A8B6E0}{0.054} & \colorbox[HTML]{E0EAF8}{0.063} & \colorbox[HTML]{E0EAF8}{0.063} & 5.169 & 5.474 & 3.996 & 4.109 & 4.102 \\ \cmidrule(lr){1-11}
    Speed Dating & \colorbox[HTML]{A8B6E0}{0.028} & \colorbox[HTML]{A8B6E0}{0.028} & \colorbox[HTML]{A8B6E0}{0.028} & \colorbox[HTML]{A8B6E0}{0.028} & \colorbox[HTML]{A8B6E0}{0.028} & 4.104 & 4.107 & 4.047 & 4.414 & 4.095 \\ \cmidrule(lr){1-11}
    Telco Customer Churn & \colorbox[HTML]{E0EAF8}{0.033} & \colorbox[HTML]{E0EAF8}{0.033} & \colorbox[HTML]{A8B6E0}{0.029} & \colorbox[HTML]{E0EAF8}{0.033} & \colorbox[HTML]{E0EAF8}{0.033} & 5.689 & 5.68 & 5.358 & 5.718 & 5.607 \\ \cmidrule(lr){1-11}
    Tic Tac Toe & 0.037 & 0.036 & \colorbox[HTML]{A8B6E0}{0.027} & 0.032 & 0.032 & 0.032 & 0.032 & \colorbox[HTML]{A8B6E0}{0.027} & 0.033 & \colorbox[HTML]{E0EAF8}{0.031} \\ \cmidrule(lr){1-11}
    Vehicle sensIT & \colorbox[HTML]{A8B6E0}{0.096} & \colorbox[HTML]{E0EAF8}{0.098} & \colorbox[HTML]{E0EAF8}{0.098} & 0.109 & 0.1 & 0.199 & 0.205 & 0.202 & 0.221 & 0.201 \\ \cmidrule(lr){1-11}
    Waveform-5000 & 0.185 & \colorbox[HTML]{E0EAF8}{0.178} & 0.188 & \colorbox[HTML]{A8B6E0}{0.176} & 0.18 & 0.399 & 0.392 & 0.396 & 0.371 & 0.364 \\ \cmidrule(lr){1-11}
    \rowcolor[HTML]{EFEFEF}
    Average rank & 3.75 & 3.333 & \colorbox[HTML]{A8B6E0}{2.083} & \colorbox[HTML]{E0EAF8}{3.25} & 3.3 & 8.483 & 8.25 & 6.65 & 8.45 & 7.45 \\ \bottomrule

\end{tabular}

\label{table:experiments}
\end{threeparttable}
\end{adjustbox}
\end{table*}

\begin{figure}[htbp]
  \centering
  \begin{minipage}[t]{0.48\linewidth}
    \centering
    \includegraphics[width=\linewidth]{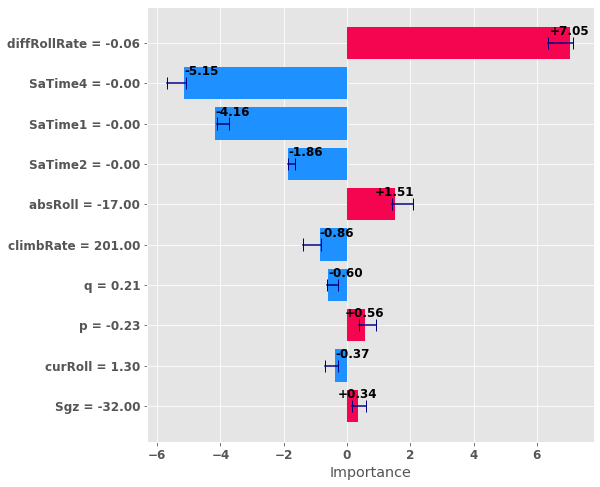}
    \caption{XGBoost with Min. Dist.}
    \label{fig:xgb_min}
    \vspace{0.5cm}
    \includegraphics[width=\linewidth]{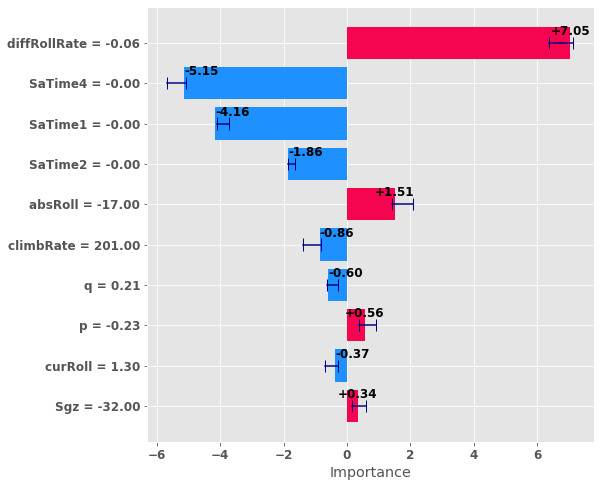}
    \caption{XGBoost with Avg. Dist.}
    \vspace{0.5cm}
    \includegraphics[width=\linewidth]{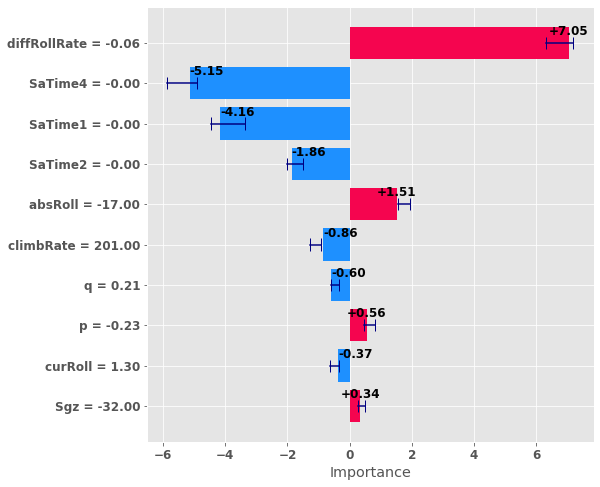}
    \caption{XGBoost with KNN}
  \end{minipage}
  \hspace{0.2cm}
  \begin{minipage}[t]{0.48\linewidth}
    \centering
    \includegraphics[width=\linewidth]{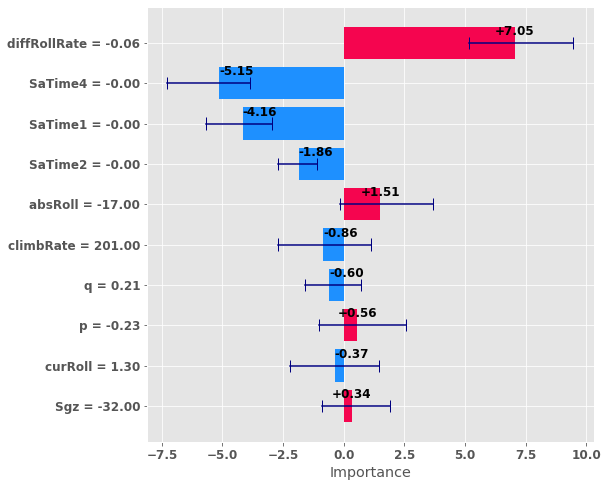}
    \caption{MLP with Min. Dist.}
    \vspace{0.5cm}
    \includegraphics[width=\linewidth]{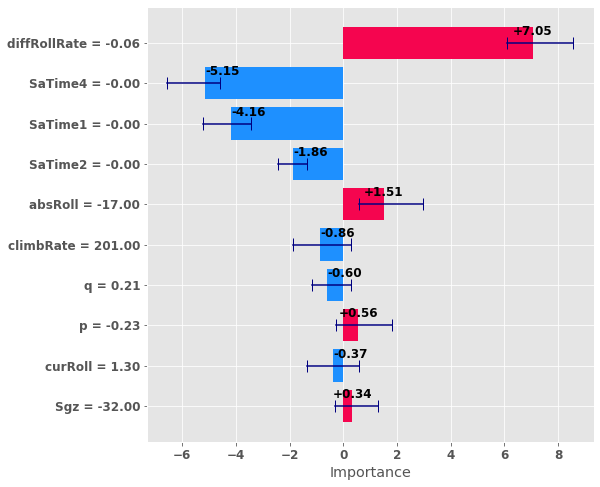}
    \caption{MLP with Avg. Dist.}
    \vspace{0.5cm}
    \includegraphics[width=\linewidth]{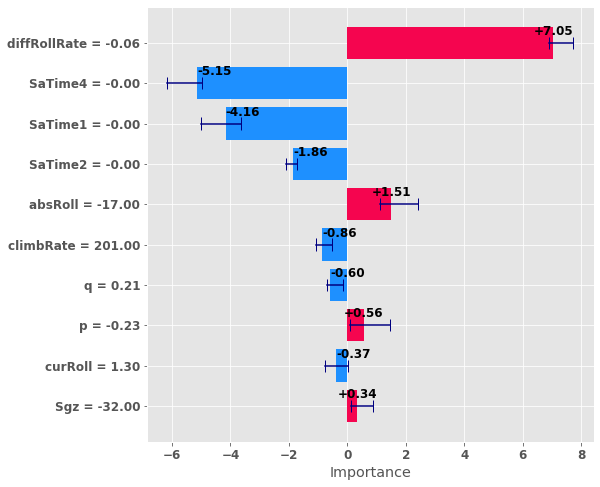}
    \caption{MLP with KNN}
  \end{minipage}
\end{figure}

\acks{This work was partially supported by the Wallenberg AI, Autonomous Systems and Software Program (WASP) funded by the Knut and Alice Wallenberg Foundation.}

\clearpage
\bibliography{references}

\appendix
\section{First Appendix} \label{appendix}

\begin{table*}[htbp]
\centering
\begin{adjustbox}{angle=0, width=1.0\textwidth}
\small
\begin{threeparttable}
\caption{The average confidence interval sizes using the \textbf{top 10} important features. The generated intervals cover the true importance scores, as predicted by the underlying explainer, with 0.95 confidence. The compared algorithms are XGBoost and MLP, with different non-conformity measures. The best-performing model is \colorbox[HTML]{A8B6E0}{colored in blue}, and the second best-performing is \colorbox[HTML]{E0EAF8}{colored in light blue}.
}
\begin{tabular}{l c c c c c | c c c c c}
    \toprule
    \rowcolor[HTML]{EFEFEF} 
    \multicolumn{1}{c}{\cellcolor[HTML]{EFEFEF}}                          & \multicolumn{5}{c}{\cellcolor[HTML]{EFEFEF}XGBoost} & \multicolumn{5}{c}{\cellcolor[HTML]{EFEFEF}Multi-Layer Perceptron} \\ \cline{2-11} 
    \rowcolor[HTML]{EFEFEF} 
    \multicolumn{1}{c}{\multirow{-2}{*}{\cellcolor[HTML]{EFEFEF}Dataset}} & Min. Dist. & Avg. Dist. & KNN & Pred. Conf. & Baseline & Min. Dist. & Avg. Dist. & KNN & Pred. Conf. & Baseline \\
    \cmidrule(lr){1-11}
    Abalone & 0.219 & 0.116 & \colorbox[HTML]{A8B6E0}{0.075} & \colorbox[HTML]{E0EAF8}{0.089} & 0.094 & 0.528 & 0.304 & 0.174 & 0.223 & 0.235 \\ \cmidrule(lr){1-11}
    Ada Prior & \colorbox[HTML]{A8B6E0}{0.087} & \colorbox[HTML]{E0EAF8}{0.088} & \colorbox[HTML]{A8B6E0}{0.087} & 0.094 & \colorbox[HTML]{E0EAF8}{0.088} & 0.232 & 0.227 & 0.243 & 0.256 & 0.228 \\ \cmidrule(lr){1-11}
    Adult & \colorbox[HTML]{E0EAF8}{0.04} & \colorbox[HTML]{E0EAF8}{0.04} & \colorbox[HTML]{A8B6E0}{0.039} & 0.043 & \colorbox[HTML]{E0EAF8}{0.04} & 0.101 & 0.099 & 0.1 & 0.112 & 0.098 \\ \cmidrule(lr){1-11}
    Bank 32 nh & \colorbox[HTML]{E0EAF8}{0.17} & \colorbox[HTML]{E0EAF8}{0.17} & 0.197 & \colorbox[HTML]{A8B6E0}{0.169} & \colorbox[HTML]{E0EAF8}{0.17} & 0.265 & 0.265 & 0.307 & 0.281 & 0.264 \\ \cmidrule(lr){1-11}
    Breast Cancer & \colorbox[HTML]{E0EAF8}{0.054} & 0.07 & \colorbox[HTML]{A8B6E0}{0.05} & 0.065 & 0.066 & 0.147 & 0.195 & 0.097 & 0.126 & 0.128 \\ \cmidrule(lr){1-11}
    Churn & 0.097 & 0.096 & 0.102 & \colorbox[HTML]{A8B6E0}{0.09} & \colorbox[HTML]{E0EAF8}{0.095} & 0.196 & 0.196 & 0.225 & 0.197 & 0.197 \\ \cmidrule(lr){1-11}
    Credit Card Fraud & \colorbox[HTML]{E0EAF8}{0.061} & \colorbox[HTML]{E0EAF8}{0.061} & 0.066 & \colorbox[HTML]{A8B6E0}{0.06} & \colorbox[HTML]{A8B6E0}{0.06} & 0.27 & 0.266 & 0.204 & 0.18 & 0.18 \\ \cmidrule(lr){1-11}
    Delta Ailerons & 0.153 & 0.067 & \colorbox[HTML]{A8B6E0}{0.054} & \colorbox[HTML]{E0EAF8}{0.061} & 0.066 & 0.175 & 0.144 & 0.137 & 0.133 & 0.141 \\ \cmidrule(lr){1-11}
    Delta Elevators & 0.082 & 0.08 & \colorbox[HTML]{A8B6E0}{0.073} & \colorbox[HTML]{E0EAF8}{0.079} & 0.08 & 0.154 & 0.151 & 0.152 & 0.158 & 0.154 \\ \cmidrule(lr){1-11}
    Electricity & \colorbox[HTML]{A8B6E0}{0.12} & \colorbox[HTML]{A8B6E0}{0.12} & 0.125 & \colorbox[HTML]{E0EAF8}{0.121} & \colorbox[HTML]{A8B6E0}{0.12} & 0.203 & 0.203 & 0.22 & 0.208 & 0.203 \\ \cmidrule(lr){1-11}
    Elevators & \colorbox[HTML]{E0EAF8}{0.07} & \colorbox[HTML]{A8B6E0}{0.069} & 0.073 & \colorbox[HTML]{E0EAF8}{0.07} & \colorbox[HTML]{A8B6E0}{0.069} & 0.156 & 0.143 & 0.144 & 0.139 & 0.13 \\ \cmidrule(lr){1-11}
    Higgs & \colorbox[HTML]{A8B6E0}{0.065} & \colorbox[HTML]{A8B6E0}{0.065} & 0.073 & \colorbox[HTML]{E0EAF8}{0.068} & \colorbox[HTML]{A8B6E0}{0.065} & 0.101 & 0.101 & 0.114 & 0.107 & 0.1 \\ \cmidrule(lr){1-11}
    JM1 & 0.14 & 0.141 & \colorbox[HTML]{A8B6E0}{0.12} & \colorbox[HTML]{E0EAF8}{0.134} & 0.141 & 0.424 & 0.265 & 0.219 & 0.227 & 0.228 \\ \cmidrule(lr){1-11}
    Madelon & 0.247 & 0.245 & 0.266 & \colorbox[HTML]{A8B6E0}{0.23} & \colorbox[HTML]{E0EAF8}{0.242} & 0.602 & 0.604 & 0.588 & 0.592 & 0.588 \\ \cmidrule(lr){1-11}
    Magic Telescope & \colorbox[HTML]{A8B6E0}{0.1} & \colorbox[HTML]{A8B6E0}{0.1} & 0.107 & 0.105 & \colorbox[HTML]{E0EAF8}{0.101} & 0.271 & 0.209 & 0.195 & 0.182 & 0.172 \\ \cmidrule(lr){1-11}
    Mozilla4 & \colorbox[HTML]{E0EAF8}{0.03} & 0.036 & \colorbox[HTML]{A8B6E0}{0.02} & 0.036 & 0.035 & 0.093 & 0.098 & 0.068 & 0.117 & 0.118 \\ \cmidrule(lr){1-11}
    MC1 & 0.035 & 0.03 & \colorbox[HTML]{A8B6E0}{0.02} & \colorbox[HTML]{E0EAF8}{0.029} & 0.03 & 0.123 & 0.13 & 0.091 & 0.142 & 0.142 \\ \cmidrule(lr){1-11}
    Numerai28.6 & \colorbox[HTML]{E0EAF8}{0.071} & 0.072 & \colorbox[HTML]{A8B6E0}{0.069} & 0.074 & 0.073 & 0.129 & 0.131 & 0.124 & 0.13 & 0.129 \\ \cmidrule(lr){1-11}
    PC2 & 0.021 & 0.021 & \colorbox[HTML]{A8B6E0}{0.015} & \colorbox[HTML]{E0EAF8}{0.02} & 0.021 & 0.314 & 0.291 & 0.165 & 0.293 & 0.295\\ \cmidrule(lr){1-11}
    Phishing & 0.049 & 0.048 & \colorbox[HTML]{A8B6E0}{0.043} & 0.047 & \colorbox[HTML]{E0EAF8}{0.046} & 0.059 & 0.059 & 0.057 & 0.061 & 0.059 \\ \cmidrule(lr){1-11}
    Phonemes & 0.176 & 0.172 & 0.182 & \colorbox[HTML]{A8B6E0}{0.16} & \colorbox[HTML]{E0EAF8}{0.166} & 0.296 & 0.258 & 0.281 & 0.252 & 0.245 \\ \cmidrule(lr){1-11}
    Pollen & \colorbox[HTML]{E0EAF8}{0.11} & 0.111 & \colorbox[HTML]{A8B6E0}{0.096} & 0.116 & 0.114 & 0.198 & 0.2 & 0.168 & 0.221 & 0.212 \\ \cmidrule(lr){1-11}
    Satellite & 0.026 & 0.025 & \colorbox[HTML]{A8B6E0}{0.017} & \colorbox[HTML]{E0EAF8}{0.024} & \colorbox[HTML]{E0EAF8}{0.024} & 0.459 & 0.374 & 0.209 & 0.292 & 0.293 \\ \cmidrule(lr){1-11}
    Scene & 0.182 & 0.177 & \colorbox[HTML]{A8B6E0}{0.149} & \colorbox[HTML]{E0EAF8}{0.16} & 0.165 & 0.478 & 0.462 & 0.422 & 0.432 & 0.433 \\ \cmidrule(lr){1-11}
    Spambase & 0.104 & 0.103 & \colorbox[HTML]{A8B6E0}{0.1} & \colorbox[HTML]{E0EAF8}{0.101} & 0.102 & 0.439 & 0.44 & 0.349 & 0.325 & 0.325 \\ \cmidrule(lr){1-11}
    Speed Dating & \colorbox[HTML]{A8B6E0}{0.113} & \colorbox[HTML]{A8B6E0}{0.113} & \colorbox[HTML]{E0EAF8}{0.127} & \colorbox[HTML]{A8B6E0}{0.113} & \colorbox[HTML]{A8B6E0}{0.113} & 0.319 & 0.318 & 0.366 & 0.322 & 0.316 \\ \cmidrule(lr){1-11}
    Telco Customer Churn & \colorbox[HTML]{E0EAF8}{0.066} & \colorbox[HTML]{E0EAF8}{0.066} & \colorbox[HTML]{A8B6E0}{0.057} & \colorbox[HTML]{E0EAF8}{0.066} & \colorbox[HTML]{E0EAF8}{0.066} & 0.16 & 0.162 & 0.139 & 0.159 & 0.159 \\ \cmidrule(lr){1-11}
    Tic Tac Toe & 0.04 & 0.039 & 0.029 & 0.034 & 0.035 & \colorbox[HTML]{E0EAF8}{0.028} & \colorbox[HTML]{E0EAF8}{0.028} & \colorbox[HTML]{A8B6E0}{0.024} & 0.03 & \colorbox[HTML]{E0EAF8}{0.028} \\ \cmidrule(lr){1-11}
    Vehicle sensIT & \colorbox[HTML]{A8B6E0}{0.09} & \colorbox[HTML]{E0EAF8}{0.091} & 0.093 & 0.104 & 0.093 & 0.159 & 0.168 & 0.168 & 0.187 & 0.165 \\ \cmidrule(lr){1-11}
    Waveform-5000 & 0.136 & \colorbox[HTML]{E0EAF8}{0.129} & 0.14 & \colorbox[HTML]{A8B6E0}{0.126} & \colorbox[HTML]{E0EAF8}{0.129} & 0.218 & 0.214 & 0.215 & 0.2 & 0.195 \\ \cmidrule(lr){1-11}
    \rowcolor[HTML]{EFEFEF}
    Average rank & 3.77 & 3.45 & \colorbox[HTML]{A8B6E0}{2.7} & \colorbox[HTML]{E0EAF8}{2.97} & 3.08 & 8.45 & 8.03 & 7.27 & 8.08 & 7.2 \\ \bottomrule
\end{tabular}

\label{table:experiments_top10}
\end{threeparttable}
\end{adjustbox}
\end{table*}

\begin{table*}[htbp]
\centering
\begin{adjustbox}{angle=0, width=1.0\textwidth}
\small
\begin{threeparttable}
\caption{The average confidence interval sizes using the \textbf{top 5} important features. The generated intervals cover the true importance scores, as predicted by the underlying explainer, with 0.95 confidence. The compared algorithms are XGBoost and MLP, with different non-conformity measures. The best-performing model is \colorbox[HTML]{A8B6E0}{colored in blue}, and the second best-performing is \colorbox[HTML]{E0EAF8}{colored in light blue}.
}
\begin{tabular}{l c c c c c | c c c c c}
    \toprule
    \rowcolor[HTML]{EFEFEF} 
    \multicolumn{1}{c}{\cellcolor[HTML]{EFEFEF}}                          & \multicolumn{5}{c}{\cellcolor[HTML]{EFEFEF}XGBoost} & \multicolumn{5}{c}{\cellcolor[HTML]{EFEFEF}Multi-Layer Perceptron} \\ \cline{2-11} 
    \rowcolor[HTML]{EFEFEF} 
    \multicolumn{1}{c}{\multirow{-2}{*}{\cellcolor[HTML]{EFEFEF}Dataset}} & Min. Dist. & Avg. Dist. & KNN & Pred. Conf. & Baseline & Min. Dist. & Avg. Dist. & KNN & Pred. Conf. & Baseline \\
    \cmidrule(lr){1-11}
    Abalone & 0.201 & 0.115 & \colorbox[HTML]{A8B6E0}{0.069} & \colorbox[HTML]{E0EAF8}{0.085} & 0.09 & 0.491 & 0.281 & 0.142 & 0.174 & 0.184 \\ \cmidrule(lr){1-11}
    Ada Prior & \colorbox[HTML]{A8B6E0}{0.08} & \colorbox[HTML]{A8B6E0}{0.08} & 0.085 & \colorbox[HTML]{E0EAF8}{0.083} & \colorbox[HTML]{A8B6E0}{0.08} & 0.22 & 0.217 & 0.244 & 0.238 & 0.214 \\ \cmidrule(lr){1-11}
    Adult & \colorbox[HTML]{E0EAF8}{0.028} & \colorbox[HTML]{E0EAF8}{0.028} & \colorbox[HTML]{A8B6E0}{0.027} & 0.03 & \colorbox[HTML]{E0EAF8}{0.028} & 0.077 & 0.076 & 0.077 & 0.086 & 0.075 \\ \cmidrule(lr){1-11}
    Bank 32 nh & \colorbox[HTML]{A8B6E0}{0.111} & 0.115 & 0.136 & \colorbox[HTML]{E0EAF8}{0.113} & 0.115 & 0.161 & 0.161 & 0.198 & 0.172 & 0.16 \\ \cmidrule(lr){1-11}
    Breast Cancer & \colorbox[HTML]{E0EAF8}{0.051} & 0.063 & \colorbox[HTML]{A8B6E0}{0.048} & 0.057 & 0.059 & 0.149 & 0.171 & 0.091 & 0.11 & 0.112 \\ \cmidrule(lr){1-11}
    Churn & 0.105 & 0.104 & 0.104 & \colorbox[HTML]{A8B6E0}{0.098} & \colorbox[HTML]{E0EAF8}{0.101} & 0.188 & 0.188 & 0.206 & 0.19 & 0.188 \\ \cmidrule(lr){1-11}
    Credit Card Fraud & \colorbox[HTML]{E0EAF8}{0.062} & \colorbox[HTML]{E0EAF8}{0.062} & 0.067 & \colorbox[HTML]{A8B6E0}{0.061} & \colorbox[HTML]{A8B6E0}{0.061} & 0.252 & 0.248 & 0.195 & 0.17 & 0.17 \\ \cmidrule(lr){1-11}
    Delta Ailerons & 0.153 & 0.067 & \colorbox[HTML]{A8B6E0}{0.054} & \colorbox[HTML]{E0EAF8}{0.061} & 0.066 & 0.175 & 0.144 & 0.137 & 0.133 & 0.141 \\ \cmidrule(lr){1-11}
    Delta Elevators & 0.079 & 0.076 & \colorbox[HTML]{A8B6E0}{0.074} & \colorbox[HTML]{E0EAF8}{0.075} & 0.076 & 0.148 & 0.141 & 0.151 & 0.145 & 0.144 \\ \cmidrule(lr){1-11}
    Electricity & \colorbox[HTML]{A8B6E0}{0.15} & \colorbox[HTML]{A8B6E0}{0.15} & \colorbox[HTML]{E0EAF8}{0.175} & \colorbox[HTML]{A8B6E0}{0.15} & \colorbox[HTML]{A8B6E0}{0.15} & 0.253 & 0.253 & 0.293 & 0.26 & 0.253 \\ \cmidrule(lr){1-11}
    Elevators & \colorbox[HTML]{E0EAF8}{0.052} & \colorbox[HTML]{A8B6E0}{0.051} & 0.056 & \colorbox[HTML]{E0EAF8}{0.052} & \colorbox[HTML]{A8B6E0}{0.051} & 0.098 & 0.093 & 0.091 & 0.088 & 0.083 \\ \cmidrule(lr){1-11}
    Higgs & \colorbox[HTML]{A8B6E0}{0.062} & \colorbox[HTML]{A8B6E0}{0.062} & 0.07 & 0.066 & \colorbox[HTML]{E0EAF8}{0.063} & 0.103 & 0.102 & 0.117 & 0.108 & 0.101 \\ \cmidrule(lr){1-11}
    JM1 & 0.14 & 0.14 & \colorbox[HTML]{A8B6E0}{0.126} & \colorbox[HTML]{E0EAF8}{0.136} & 0.14 & 0.409 & 0.243 & 0.213 & 0.211 & 0.211 \\ \cmidrule(lr){1-11}
    Madelon & 0.226 & 0.224 & 0.242 & \colorbox[HTML]{A8B6E0}{0.213} & \colorbox[HTML]{E0EAF8}{0.222} & 0.615 & 0.617 & 0.593 & 0.608 & 0.603 \\ \cmidrule(lr){1-11}
    Magic Telescope & \colorbox[HTML]{A8B6E0}{0.073} & \colorbox[HTML]{A8B6E0}{0.073} & 0.077 & \colorbox[HTML]{E0EAF8}{0.076} & \colorbox[HTML]{A8B6E0}{0.073} & 0.159 & 0.13 & 0.119 & 0.112 & 0.106 \\ \cmidrule(lr){1-11}
    Mozilla4 & 0.035 & \colorbox[HTML]{E0EAF8}{0.034} & \colorbox[HTML]{A8B6E0}{0.019} & \colorbox[HTML]{E0EAF8}{0.034} & \colorbox[HTML]{E0EAF8}{0.034} & 0.097 & 0.101 & 0.071 & 0.121 & 0.122 \\ \cmidrule(lr){1-11}
    MC1 & \colorbox[HTML]{E0EAF8}{0.034} & 0.035 & \colorbox[HTML]{A8B6E0}{0.023} & 0.035 & 0.035 & 0.118 & 0.125 & 0.089 & 0.136 & 0.137 \\ \cmidrule(lr){1-11}
    Numerai28.6 & \colorbox[HTML]{E0EAF8}{0.07} & 0.071 & \colorbox[HTML]{A8B6E0}{0.068} & 0.073 & 0.072 & 0.126 & 0.128 & 0.121 & 0.127 & 0.126 \\ \cmidrule(lr){1-11}
    PC2 & 0.021 & \colorbox[HTML]{E0EAF8}{0.02} & \colorbox[HTML]{A8B6E0}{0.015} & \colorbox[HTML]{E0EAF8}{0.02} & 0.021 & 0.266 & 0.259 & 0.157 & 0.252 & 0.252 \\ \cmidrule(lr){1-11}
    Phishing & 0.056 & 0.055 & \colorbox[HTML]{A8B6E0}{0.05} & 0.054 & \colorbox[HTML]{E0EAF8}{0.053} & 0.057 & 0.058 & 0.056 & 0.059 & 0.058 \\ \cmidrule(lr){1-11}
    Phonemes & 0.176 & 0.172 & 0.182 & \colorbox[HTML]{A8B6E0}{0.16} & \colorbox[HTML]{E0EAF8}{0.166} & 0.296 & 0.258 & 0.281 & 0.252 & 0.245 \\ \cmidrule(lr){1-11}
    Pollen & \colorbox[HTML]{E0EAF8}{0.11} & 0.111 & \colorbox[HTML]{A8B6E0}{0.096} & 0.116 & 0.114 & 0.198 & 0.2 & 0.168 & 0.221 & 0.212 \\ \cmidrule(lr){1-11}
    Satellite & 0.027 & 0.027 & \colorbox[HTML]{A8B6E0}{0.019} & \colorbox[HTML]{E0EAF8}{0.025} & 0.026 & 0.364 & 0.323 & 0.181 & 0.264 & 0.264 \\ \cmidrule(lr){1-11}
    Scene & 0.15 & 0.143 & \colorbox[HTML]{A8B6E0}{0.102} & \colorbox[HTML]{E0EAF8}{0.131} & 0.134 & 0.277 & 0.277 & 0.22 & 0.261 & 0.256 \\ \cmidrule(lr){1-11}
    Spambase & \colorbox[HTML]{E0EAF8}{0.09} & \colorbox[HTML]{E0EAF8}{0.09} & \colorbox[HTML]{A8B6E0}{0.075} & \colorbox[HTML]{E0EAF8}{0.09} & \colorbox[HTML]{E0EAF8}{0.09} & 0.32 & 0.297 & 0.249 & 0.257 & 0.25 \\ \cmidrule(lr){1-11}
    Speed Dating & \colorbox[HTML]{E0EAF8}{0.11} & \colorbox[HTML]{E0EAF8}{0.11} & 0.127 & 0.111 & \colorbox[HTML]{A8B6E0}{0.109} & 0.312 & 0.311 & 0.358 & 0.315 & 0.308 \\ \cmidrule(lr){1-11}
    Telco Customer Churn & \colorbox[HTML]{E0EAF8}{0.062} & \colorbox[HTML]{E0EAF8}{0.062} & \colorbox[HTML]{A8B6E0}{0.054} & 0.063 & 0.063 & 0.155 & 0.156 & 0.131 & 0.158 & 0.155 \\ \cmidrule(lr){1-11}
    Tic Tac Toe & 0.037 & 0.037 & \colorbox[HTML]{E0EAF8}{0.027} & 0.031 & 0.032 & \colorbox[HTML]{E0EAF8}{0.027} & \colorbox[HTML]{E0EAF8}{0.027} & \colorbox[HTML]{A8B6E0}{0.022} & 0.029 & \colorbox[HTML]{E0EAF8}{0.027} \\ \cmidrule(lr){1-11}
    Vehicle sensIT & \colorbox[HTML]{A8B6E0}{0.071} & \colorbox[HTML]{E0EAF8}{0.072} & 0.073 & 0.082 & 0.074 & 0.119 & 0.126 & 0.123 & 0.14 & 0.122 \\ \cmidrule(lr){1-11}
    Waveform-5000 & 0.142 & \colorbox[HTML]{E0EAF8}{0.136} & 0.151 & \colorbox[HTML]{A8B6E0}{0.133} & 0.136 & 0.225 & 0.221 & 0.218 & 0.2 & 0.191 \\ \cmidrule(lr){1-11}
    \rowcolor[HTML]{EFEFEF}
    Average rank & 3.78 & 3.33 & \colorbox[HTML]{A8B6E0}{2.7} & \colorbox[HTML]{E0EAF8}{3.07} & 3.12 & 8.43 & 8.2 & 7.3 & 8.1 & 6.97 \\ \bottomrule
\end{tabular}
\label{table:experiments_top5}
\end{threeparttable}
\end{adjustbox}
\end{table*}

\begin{figure*}[ht]
    \centering
    \includegraphics[width=1.\textwidth]{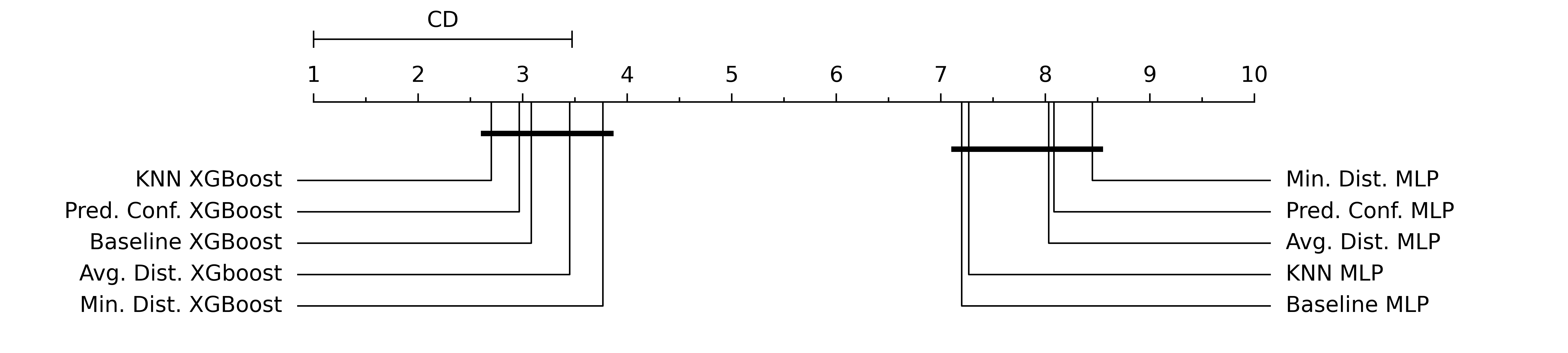}
    \caption{The average rank of the compared regression models on the 30 datasets with respect to the interval size of the \textbf{top 10} features (a lower rank is better), where the critical difference (CD) represents the largest difference that is not statistically significant.}
    \label{fig:ranks_top10}
\end{figure*}

\begin{figure*}[ht]
    \centering
    \includegraphics[width=1.\textwidth]{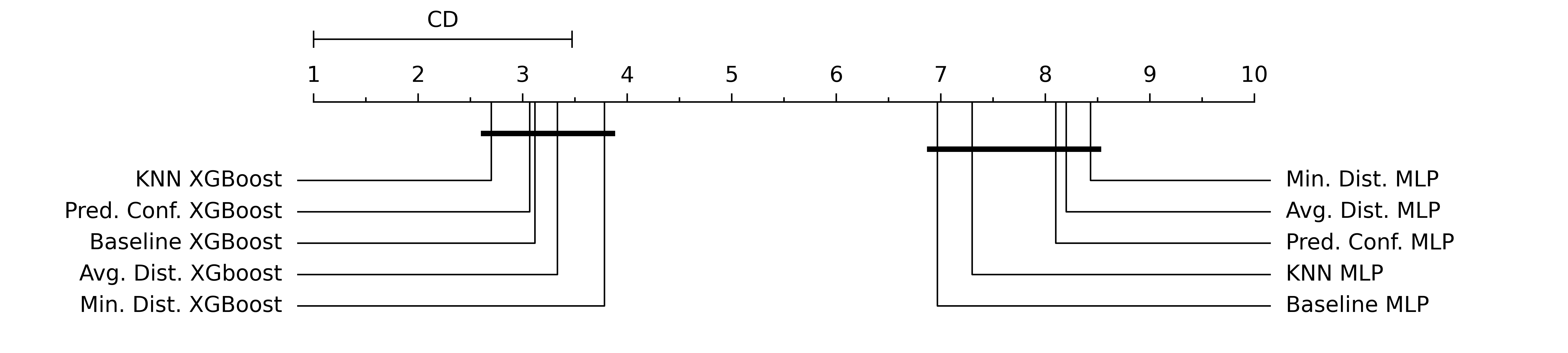}
    \caption{The average rank of the compared regression models on the 30 datasets with respect to the interval size of the \textbf{top 5} features (a lower rank is better), where the critical difference (CD) represents the largest difference that is not statistically significant.}
    \label{fig:ranks_top5}
\end{figure*}
\end{document}